# Enhancing Medical Image Registration via Appearance Adjustment Networks

Mingyuan Meng [a], Lei Bi [a, *], Michael Fulham [a, b], David Dagan Feng [a, c], and Jinman Kim [a, *]

[a] School of Computer Science, the University of Sydney, Australia.
[b] Department of Molecular Imaging, Royal Prince Alfred Hospital, Australia.
[c] Med-X Research Institute, Shanghai Jiao Tong University, Shanghai, China.

**Abstract** — Deformable image registration is fundamental for many medical image analyses. A key obstacle for accurate image registration lies in image appearance variations such as the variations in texture, intensities, and noise. These variations are readily apparent in medical images, especially in brain images where registration is frequently used. Recently, deep learning-based registration methods (DLRs), using deep neural networks, have shown computational efficiency that is several orders of magnitude faster than traditional optimization-based registration methods (ORs). DLRs rely on a globally optimized network that is trained with a set of training samples to achieve faster registration. DLRs tend, however, to disregard the target-pair-specific optimization inherent in ORs and thus have degraded adaptability to variations in testing samples. This limitation is severe for registering medical images with large appearance variations, especially since few existing DLRs explicitly take into account appearance variations. In this study, we propose an Appearance Adjustment Network (AAN) to enhance the adaptability of DLRs to appearance variations. Our AAN, when integrated into a DLR, provides appearance transformations to reduce the appearance variations during registration. In addition, we propose an anatomy-constrained loss function through which our AAN generates anatomy-preserving transformations. Our AAN has been purposely designed to be readily inserted into a wide range of DLRs and can be trained cooperatively in an unsupervised and end-to-end manner. We evaluated our AAN with three state-of-the-art DLRs - Voxelmorph (VM), Diffeomorphic Voxelmorph (DifVM), and Laplacian Pyramid Image Registration Network (LapIRN) – on three well-established public datasets of 3D brain magnetic resonance imaging (MRI) - IBSR18, Mindboggle101, and LPBA40. The results show that our AAN consistently improved existing DLRs and outperformed state-of-the-art ORs on registration accuracy, while adding a fractional computational load to existing DLRs.

## 1. Introduction

Medical image registration is fundamental for a variety of medical image analyses and has been a focus of active research for many years [1][2]. For example, image registration is necessary when analyzing a group of images that are acquired from different timepoints or different scanners [29]. For brain images, image registration is frequently used, for example, to register images with a brain atlas for localization or to register sequential images to detect disease progression. Image registration aims to establish a pixel/voxel-level correspondence between a pair of fixed and moving images/volumes. The correspondence is a spatial transformation, where the fixed and moving images/volumes can be warped so that they are spatially aligned to each other. A key obstacle for accurate image registration lies in the variations in image appearance. Similar to Zhao et al. [16], by 'appearance' we mean organ tissue texture, intensity distribution, and image noise. The variations in image appearance are readily apparent in brain images where registration is frequently used. The challenges of appearance variations have been mitigated by image preprocessing

*Published at NeuroImage*

\* Corresponding authors.
Emails address: lei.bi@sydney.edu.au (L. Bi) and jinman.kim@sydney.edu.au (J. Kim)
This work was supported in part by the Australian Research Council (ARC) grants (DP200103748).

such as image normalization, or intensity-robust loss function such as normalized cross-correlation [20], but the ambiguities in image appearances (e.g., diverse tissue texture and image noise) can still lead to inaccurate registration [16].

Early image registration methods employed key points such as scale-invariant feature transform (SIFT) [44] and speeded up robust features (SURF) [45]. In these methods, the parameters of homography matrix are calculated using matched key points in the fixed and moving images. These methods have the ability to perform image registration in real-time (less than 1 second), but they have difficulties in handling non-linear deformable spatial transformations. Unfortunately, the spatial transformations required to optimally register medical images are highly non-linear, especially for regions such as the cerebral cortex where there are ubiquitous non-linear deformations [46]. Therefore, deformable image registration has become a research focus of recent medical image registration studies [3][5][12][25][38][39][46].

For deformable medical image registration, traditional methods are based on image features/intensity and can be summarized as an optimization problem [19][20][21][30][48]. These optimization-based registration methods (ORs) generally define a hypothesis space of possible spatial transformations and define similarity metrics between fixed and moving images. Then, the maximum of the similarity metrics is found by iteratively optimizing/updating the spatial transformation. However, such pairwise (target-pair-specific) optimization is computationally expensive [3][5][7] and it may take several hours to register a pair of images with CPUs [21][23]. Recent GPU implementations of ORs has reduced the runtime to minutes but still not within the range of real-time registration [42]. Furthermore, registration accuracy of ORs is still unsatisfactory for clinical practice. Even state-of-the-art ORs, such as SyN (Symmetric Normalization method) [20] and DRAMMS (Deformable Registration via Attribute Matching and Mutual-Saliency weighting) [48], are prone to errors in difficult brain registration studies.

Deep learning-based registration methods (DLRs) have been proposed recently for medical images. Compared to traditional ORs, DLRs have higher or competitive registration accuracy. DLRs adopt an end-to-end neural network to directly predict a spatial transformation in a single pass [3][4][5][6][7]. A key advantage of DLRs is the computational efficiency that is several orders of magnitude faster than ORs. DLRs can register a pair of images within a minute using CPUs or within a second using GPUs [3][4][5][7]. DLRs achieve this by discarding the target-pair-specific optimization that is inherent in ORs and instead rely on a globally optimized network that is trained with a set of training samples. However, compared to the target-pair-specific optimization, globally optimized networks inherently have degraded adaptability to variations in testing samples. This limitation is severe for registering medical images with large appearance variations, especially since few existing DLRs explicitly take into account appearance variations. Consequently, the performance of existing DLRs degrades markedly when they are applied to unseen fixed/moving image pairs whose appearances are observably different (e.g., Fig. 1a and Fig. 1b).

In this study, we overcome this limitation through an Appearance Adjustment Network (AAN). We suggest that our method is the first to enhances DLRs for medical images via automatic appearance adjustment. Our AAN is designed to be readily inserted into a wide range of existing DLRs to reduce the appearance differences between fixed and moving images. It performs a high-quality nonlinear tissue-intensity mapping, which takes a fixed/moving image pair and an automatically-extracted anatomy edge map as the input and then predicts an appearance transformation (e.g., Fig. 1d) to adjust the moving image's appearance. Through a novel anatomy-constrained loss function based on the anatomy edge map, the AAN can make the moving image's appearance approach the fixed image's appearance while retaining its original anatomical structures (e.g., Fig. 1b and Fig. 1c). In doing so, the AAN can enhance the adaptability of DLRs to appearance variations. Our AAN also enables DLRs to focus on learning spatial transformations by suppressing the appearance variations in image pairs so that it can increase the learning efficiency of DLRs and reduce the requirement for the number of training samples.

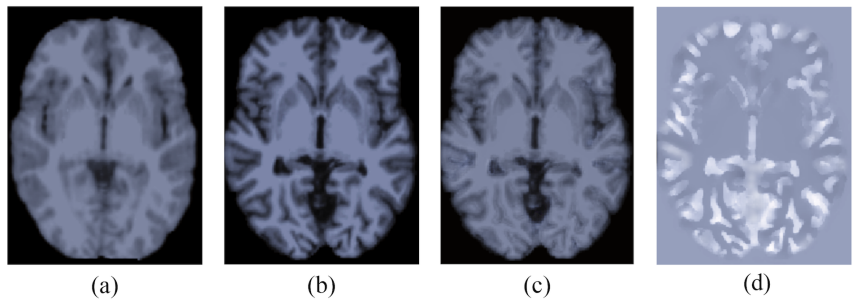

(a) (b) (c) (d)

Fig. 1. Examples of transaxial MRI slices of (a) fixed and (b) moving images. The image appearances of (a) and (b) are observably different. In (c) we show the result after appearance adjustment of (b), and (d) shows the appearance transformation between (b) and (c). The appearance transformation (d) turns (b) into (c), making (c)'s image appearance similar to (a) while preserving the anatomical structures in (b).

Our AAN was motivated by the methods proposed by Uzunova et al. [15] and Zhao et al. [16]. Uzunova et al. [15] leveraged existing appearance variations and built a statistical appearance model from a few training samples to perform data augmentation for DLRs. Zhao et al. [16] trained two separate convolutional neural networks (CNNs) to estimate spatial and appearance transformations between a fixed atlas image and other images. Then, cross-combinations derived from the estimated spatial and appearance transformations are applied to the fixed atlas image to synthesize new training samples for deep learning-based medical image segmentation. We also use a neural network (i.e., AAN) to estimate the appearance transformations between the fixed and moving images and assemble it with existing DLRs. However, in contrast to the studies of Uzunova et al. [15] and Zhao et al. [16] that involved laborious augmentation of the variations in training samples to improve the network's adaptability to the appearance variations in unseen testing samples, our AAN reduces the appearance variations in both training and testing samples, which enables DLRs to focus on learning spatial transformations.

We evaluated our AAN with three state-of-the-art DLRs - Voxelmorph (VM) [3], Diffeomorphic Voxelmorph (DifVM) [12], and Laplacian Pyramid Image Registration Network (LapIRN) [55] - on three well-benchmarked datasets of 3D brain magnetic resonance imaging (MRI) - IBSR18 [8], Mindboggle101 [9], and LPBA40 [28]. The experimental results show that our AAN can generate consistent improvements in registration accuracy and training efficiency.

## 2. Related Work
### 2.1. Deep Learning-based Medical Image Registration

Existing DLRs use deep neural networks to directly learn a complex mapping from a pair of fixed/moving images to a spatial transformation. Supervised DLRs trained a network using pairs of fixed/moving images as input and ground truth transformations as labels [6][10][11][25][37][38]. Ground truth transformations, however, are usually difficult to acquire, and thus flawed transformations measured by traditional ORs were used as ground truth labels in some studies [5][10][25][38], which introduced inherited registration errors. Alternatively, synthetic image pairs with ground truth transformations can be used, but the synthetic data have to be carefully designed so as to resemble the real data [40].

Weakly-supervised DLRs removed the reliance on ground truth transformations by incorporating manually depicted anatomy labels and have shown improved performance in registration accuracy [13][14][22][41]. The manual derivation of anatomy labels, however, is time-consuming. Besides, weakly-supervised DLRs train a network based on the similarity between the anatomy masks of fixed and moving images, where the anatomy mask of the moving image is warped to be spatially aligned to the fixed image's mask. This means that the network learns to align the labeled regions rather than to register all image pixels, and hence exposing the pixels in the periphery of the labeled regions to registration errors [3]. In addition, similar to supervised DLRs, the registration accuracy of weakly-supervised DLRs is still bounded by the quality of anatomy labels.

Unsupervised DLRs have been proposed to eliminate the reliance of DLRs on label information [4][7][12][39][40]. Unsupervised DLRs train a network to minimize dissimilarity metrics such as intensity Mean Squared Error (MSE) between the fixed and moving images. Unsupervised DLRs are considered more promising, compared to supervised counterparts, because synthetic or inaccurate label information are not incorporated into the network's training. Recently, unsupervised DLRs tend to decouple the complex registration process into multiple steps, and this has been demonstrated to improve the registration accuracy [54][55]. For example, Hu et al. [54] proposed a Dual-stream Pyramid Registration Network (Dual-PRNet) that decouples feature learning from transformation optimization. Mok et al. [55] proposed a Laplacian Pyramid Image Registration Network (LapIRN), which decouples the registration process with 3-level pyramid networks and achieved the top-performance at the 2021 Learn2Reg challenge [56].

The limitations associated with the availability of ground truth labels have been largely mitigated by unsupervised DLRs, but DLRs retain another key limitation in their reliance on using globally optimized networks. As mentioned above, globally optimized networks inherently have degraded adaptability to variations in unseen testing samples. In some studies [3][14], test-specific iterative refinement was incorporated into DLRs to overcome this limitation, where the trained network is fine-tuned individually for each testing image pair. It is worth noting that iteratively-refined DLRs performed better than non-iterative (one-pass) counterparts but at a higher computational complexity, thus reducing the advantage of DLRs on computational efficiency.

In this study, in contrast to using test-specific iterative refinements, we propose an AAN to enhance the adaptability of DLRs to appearance variations. We focus on inserting the AAN into non-iterative unsupervised DLRs that remove the reliance on using test-specific refinements and data labels for training.

## 2.2. Appearance Variations

Although many investigators have focused on the DLR's network architectures, loss functions, and learning schemes, there are few studies that have investigated on the appearance variations that are apparent in medical images. Normalized cross-correlation, an intensity-robust loss function, is considered more robust to appearance variations [20], but it is unable to completely resolve appearance variations as demonstrated in our experiments (Section 5). Traditional histogram matching/equalization is considered to be a helpful preprocessing step to reduce appearance variations [31][32]. However, these methods are unable to handle the variations beyond the histogram differences such as the variations in tissue texture and image noise. In addition, data augmentation can improve the network's adaptability to appearance variations [15][16], but the generation of synthetic data is time-consuming and has to be carefully designed so that the synthetic data could have realistic appearance.

Metamorphosis is a concept that can be applied to image registration. Traditional metamorphosis methods jointly estimate a space deformation and an appearance change to construct a spatio-temporal trajectory that morphs a source image to a target image, where appearance changes are modeled as transformation variables via image similarity metrics [49][50]. Recently, François et al. [51] proposed combining metamorphosis and Large Deformation Diffeomorphic Metric Mapping (LDDMM) for optimization-based image registration. Bône et al. [52] introduced metamorphosis into deep learning and proposed a metamorphic auto-encoder for shape analysis, which jointly learns shape and appearance representations. Metamorphosis, however, has yet to be introduced to DLRs. Although we have mentioned that some DLRs decoupled registration to some degree [54][55] (Section 2.1), few DLRs decouple appearance transformations from image registration. Our AAN follows a similar concept to metamorphosis, where we model the appearance transformations between the fixed and moving images to enhance DLRs. It should be noted, nevertheless, that we still focus on image registration where moving images are warped (not morphed) into fixed images. Therefore, the appearance transformations (produced by the AAN) are used only to reduce appearance variations, enabling DLRs to obtain more accurate spatial transformations, but they are not used to transform moving images as final registration results.

A similar architecture of inserting Image Transformer Networks (ITNs) into Spatial Transformer Networks (STNs) was proposed by Lee et al. [14]. In their study, the ITNs extracted Structures-of-Interests (SoIs), such as segmentation or landmarks, and removed unlabeled regions, thereby markedly narrowing the registration space for the STNs. Therefore, the registration performance is reliant on the availability of SoI labels. Our AAN, in contrast, reduces appearance variations by adjusting image appearance, through which we further leverage the information in image appearance without the need of supplementary information offered by anatomy labels.

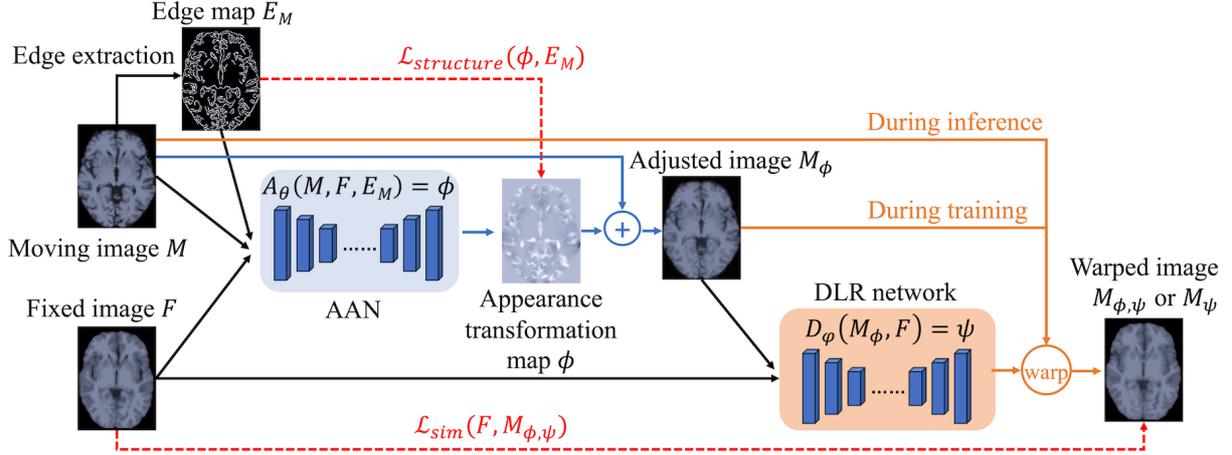

Fig. 2. The workflow of AAN-enhanced DLRs for 3D brain MRI registration. The proposed AAN is in the blue rectangle. The orange rectangle includes an existing DLR network. The whole architecture (including $A_\theta$ and $D_\varphi$) is end-to-end trained to maximize the similarity between $F$ and $M_{\phi,\psi}$ in an unsupervised manner. The $L_{structure}(\phi, E_M)$ is an anatomy-constrained loss which penalizes the $\phi$ that alters the $M$'s original anatomical structures.

## 3. Method

### 3.1. Overview

Let moving image $M$ and fixed image $F$ be two volumes defined over a 3D spatial domain $\Omega \subset \mathbb{R}^3$. Image registration aims to find a spatial transformation $\psi$ that warps the moving image $M$, so that the warped image $M \circ \psi$ is spatially aligned with the fixed image $F$. Traditional ORs usually formulate registration as an optimization problem that aims to find the most optimized $\psi$ to minimize the energy:

$$\psi = \underset{\psi}{\mathrm{argmin}}\, \mathcal{L}_{similarity}(F, M \circ \psi) + \mu \mathcal{L}_{smooth}(\psi), \qquad (1)$$

where the first term $\mathcal{L}_{similarity}(F, M \circ \psi)$ measures the image difference between fixed image and warped image, the second term $\mathcal{L}_{smooth}(\psi)$ imposes regularization on $\psi$, and $\mu$ is a regularization parameter.

DLRs use a neural network (named DLR network) to model a function that directly predicts the spatial transformation $\psi$:

$$D_\varphi(M, F) = \psi, \qquad (2)$$

where $\varphi$ denotes the network's learnable parameters, through which DLRs replace the target-pair-specific optimization by global optimization of parameters $\varphi$.

Fig. 2 outlines the workflow of AAN-enhanced DLRs for 3D brain MRI registration. Our AAN (blue rectangle in Fig. 2) can be assembled with a variety of existing DLR networks (orange rectangle in Fig. 2) that can receive two images and then predict a spatial transformation $\psi$ between them. The whole workflow only takes the moving image $M$ and fixed image $F$ as input.

Initially, the anatomy edges of the image $M$ are automatically extracted into a binary edge map $E_M$ (1 for edges and otherwise 0). The edge map can provide anatomical information and assist in producing anatomy-preserving appearance transformation. The $E_M$ is directly fed into the AAN as supplementary input and is also used to supervise the AAN's training through an anatomy-constrained loss term $L_{structure}$ (see Section 3.3.1). Since our AAN doesn't require accurate anatomy edge maps, 3D Canny edge detection [36] is adopted.

Then, a neural network (i.e., AAN) is used to model a function:

$$A_\theta(M, F, E_M) = \phi, \qquad (3)$$

where $\phi$ is an appearance transformation, $\theta$ are the learnable parameters of AAN, and $M, F, E_M$ are concatenated as the input of $A_\theta$. This function $A_\theta$ is a nonlinear tissue-intensity mapping from images $M, F$, and edge map $E_M$ to an appearance transformation $\phi$. Note that $E_M$ is used as supplementary input to provide anatomical information to $A_\theta$, which helps the AAN to produce anatomy-preserving transformations. The $\phi$ is voxel-wise added to $M$, resulting in an appearance adjusted moving image $M_\phi = M + \phi$.

Finally, $M_\phi$ and $F$ are fed into a DLR network which is denoted by $D_\varphi(M_\phi, F) = \psi$. During training, the appearance adjusted image $M_\phi$ is warped to be $M_{\phi,\psi} = M_\phi \circ \psi$, and the whole architecture (including $A_\theta$ and $D_\varphi$) is trained end-to-end to maximize the similarity between $F$ and $M_{\phi,\psi}$ in an unsupervised manner. However, during inference, the unadjusted moving image $M$ is warped to be the final registration result $M_\psi = M \circ \psi$. Note that, although the $M_\phi$ is not warped during inference, the AAN is still used to produce the $M_\phi$ as the $\psi$ is always predicted based on the $M_\phi$ and $F$.

### 3.2. AAN Architecture

The function $A_\theta(M, F, E_M) = \phi$ is modeled based on a CNN architecture similar to U-net [17], which consists of encoder and decoder stages with skip connections. Fig. 3 depicts the AAN architecture used in our experiments. $M, F$, and $E_M$ are concatenated and used as the input. The output is a 1-channel 3D volume $\phi$ in the same resolution as the input volume. We apply 3D convolutions in the encoder and decoder stages using a kernel size of $3 \times 3 \times 3$. Except for the last convolution layer, each other convolution layer is followed by a LeakyReLU layer with a parameter of 0.2.

In the encoder stage, we use convolutions with a stride of 2 to reduce the volume resolution by half at each layer. At each strided convolution step, we double the number of feature channels. Each successive convolution layer of the encoder operates over coarser representations of the input and handles larger-scale features.

In the decoder stage, we use upsampling layers to upsample the volume resolution, followed by convolutions with a stride of 1 to reduce the number of feature channels. To recover the features lost during the strided convolution process, skip connections are used to concatenate the low-level features from the encoder stage and the upsampled high-level features in the decoder stage. At the final layer, a convolution layer is used to map 8-channel feature maps to the desired appearance transformation $\phi$.

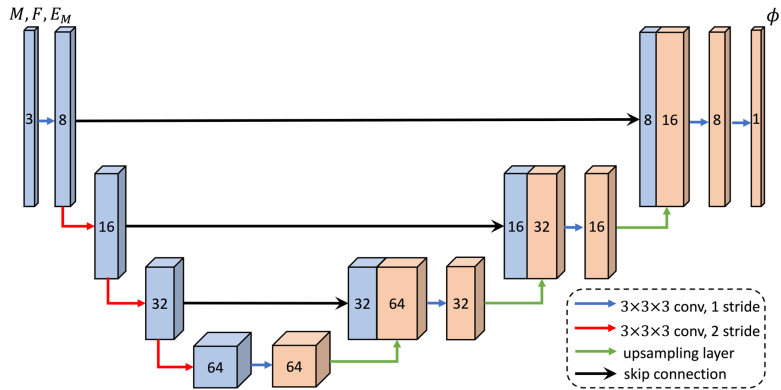

Fig. 3. The AAN architecture used in our experiments. Each cube represents 3D volumes (feature maps), generated from the preceding volumes using a 3D convolutional layer. The channel number of each volume is shown within the cubes. Skip connection is used to concatenate encoder and decoder features.

### 3.3. Loss Function

We propose an anatomy-constrained loss term $L_{structure}$ to penalize appearance transformations that alter an image's original anatomical structures, enabling AAN to produce anatomy-preserving appearance transformation. Then, we describe the DLR loss functions used in our experiments.

### 3.3.1. AAN Loss Function

During appearance adjustment, the original anatomical structures of the moving image have to be preserved, or the spatial transformation $\psi$ predicted based on the adjusted image $M_\phi$ and fixed image $F$ cannot be used to register the original moving image $M$ with the fixed image $F$. So when the AAN is inserted into a DLR, an extra loss term $\mathcal{L}_{structure}$ needs to be added to penalize appearance transformations that alter an image's original anatomical structures. $\mathcal{L}_{structure}$ is defined as:

$$L_{structure}(\phi, E_M) = \sum_{p \in \Omega} [1 - E_M(p)] \, ||\nabla \phi(p)||, \qquad (4)$$

where $E_M$ is the binary edge map of $M$ (1 for edges and otherwise 0) and $\nabla$ is the spatial gradient operator. $\mathcal{L}_{structure}$ calculates the sum of $||\nabla \phi(p)||$ at the location $p$ where $E_M(p) = 0$. Intuitively, this loss penalizes unsmooth $\phi$ within each anatomical structure but allows for discontinuities at anatomical boundaries. After $\mathcal{L}_{structure}$ is added, the total loss function $\mathcal{L}_{total}$ is:

$$\mathcal{L}_{total} = \mathcal{L}_{DLR} + \lambda \mathcal{L}_{structure}, \qquad (5)$$

where $\mathcal{L}_{DLR}$ is the loss function of the DLR network, and $\lambda$ is a regularization parameter of $\mathcal{L}_{structure}$.

### 3.3.2. DLR Loss Function

For unsupervised DLRs, $\mathcal{L}_{DLR}$ evaluates the model using only the input volumes and the predicted registration results. An unsupervised $\mathcal{L}_{DLR}$ usually consists of a loss term $\mathcal{L}_{sim}$ that penalizes the differences between warped image $M_{\phi,\psi}$ and fixed image $F$, and at least one regularization term $R(\psi)$ that penalizes physically unrealistic $\psi$:

$$\mathcal{L}_{DLR} = \mathcal{L}_{sim}(F, M_{\phi,\psi}) + \mu_1 R_1(\psi) + \cdots + \mu_n R_n(\psi), \qquad (6)$$

where $R_1, R_2, \ldots, R_n$ are regularization terms, and $\mu_1, \mu_2, \ldots, \mu_n$ are their regularization parameters. Since $M_{\phi,\psi} = M_\phi \circ \psi = (M + \phi) \circ \psi$, both the AAN and the DLR network are trained cooperatively to minimize $\mathcal{L}_{sim}$. Note that a variety of $\mathcal{L}_{DLR}$ may work similarly well and that the exact $\mathcal{L}_{DLR}$ is not our research focus. Supervised or weakly-supervised DLR networks can also be considered if we add extra loss terms into $\mathcal{L}_{DLR}$ such as the MSE between ground truth transformations and the predicted $\psi$, or a similarity term that penalizes the differences between fixed and warped anatomy masks.

We experimented with two widely used loss functions for $\mathcal{L}_{sim}$. The first is the intensity MSE between $F$ and $M_{\phi,\psi}$:

$$MSE(F, M_{\phi,\psi}) = \frac{1}{|\Omega|} \sum_{p \in \Omega} [F(p) - M_{\phi,\psi}(p)]^2. \qquad (7)$$

The second is the Local Cross-Correlation (LCC) of $F$ and $M_{\phi,\psi}$, which is considered robust to appearance variations among MRI scans [20]. Following [3], Let $\hat{I}(p)$ denote the local mean intensity of image $I$ in the location $p$:

$$\hat{I}(p) = \frac{1}{n^3} \sum_{p_i} I(p_i), \qquad (8)$$

where $p_i$ iterates over a $n^3$ neighboring volume around $p$, with $n = 9$ in our experiments. The LCC of $F$ and $M_{\phi,\psi}$ is defined as:

$$LCC(F, M_{\phi,\psi}) = -\sum_{p \in \Omega} \frac{\left(\sum_{p_i} [F(p_i) - \hat{F}(p)][M_{\phi,\psi}(p_i) - \hat{M}_{\phi,\psi}(p)]\right)^2}{\left(\sum_{p_i} [F(p_i) - \hat{F}(p)]^2\right)\left(\sum_{p_i} [M_{\phi,\psi}(p_i) - \hat{M}_{\phi,\psi}(p)]^2\right)}. \qquad (9)$$

Minimizing $\mathcal{L}_{sim}$ encourages $M_{\phi,\psi}$ to approximate $F$ but may generate a non-smooth or non-invertible $\psi$ that is not physically realistic. Therefore, some regularization terms are needed to penalize local spatial variations [7] or negative Jacobian determinants [46]. In our experiments, we directly used the default regularization settings of baseline DLRs if no extra setting is stated.

## 4. Experimental Setup

We evaluated our AAN with well-established public 3D brain MRI datasets, which have been widely used for evaluation in various registration studies [3][5][7][10][12][14][46]. We inserted our AAN into existing DLRs and compare the AAN-enhanced DLRs with traditional ORs.

## 4.1. Datasets

Our training set consists of 2,760 T1–weighted brain MRI volumes acquired from four publicly available datasets:

- ADNI [33]: The Alzheimer's Disease Neuroimaging Initiative (ADNI) project includes MRI scans from healthy subjects or Alzheimer Disease (AD) patients. From ADNI, 121 MRI scans were used.
- ABIDE [34]: The Autism Brain Imaging Data Exchange (ABIDE) involves 1,112 subjects with MRI scans. From ABIDE, 1,088 MRI scans were used.
- ADHD [35]: The ADHD-200 Sample is a grassroots initiative, dedicated to accelerating the scientific community's understanding of the neural basis of Attention Deficit Hyperactivity Disorder. From ASHD-200, 970 MRI scans were used.
- IXI [18]: Information eXtraction from Images (IXI) dataset consists of 581 MRI scans from normal, healthy subjects. All 581 scans from IXI were used.

Our validation and testing set consist of three publicly available, manually segmented, brain MRI datasets:

- IBSR18 [8]: The Internet Brain Segmentation Repository (IBSR) dataset has 18 skull-stripped T1-weighted brain MRI scans from healthy subjects, along with expert manually derived segmentation result. All 18 scans were randomly split into 6 and 12 scans for validation and testing sets.
- LPBA40 [28]: The LONI Probabilistic Brain Atlas (LPBA) dataset contains 40 skull-stripped T1-weighted brain MRI scans from healthy subjects, each of which comes with segmentation labels of 56 anatomical structures. We randomly chose 10 scans as the validation set, and the remaining 30 scans were used as the testing set.
- Mindboggle101 [9]: The Mindboggle101 dataset contains 101 skull-stripped T1-weighted brain MRI scans from healthy subjects. The latest version (v3 on 2019/04/03) has 100 scans that come with manually segmented labels. All 100 available scans were randomly split into 30 and 70 scans as the validation and testing sets.

## 4.2. Baseline Methods

Three unsupervised DLRs were used as baselines. These methods have been widely benchmarked [25][39][40][43][56]:

- Voxelmorph (VM) [3] – a U-net based network for deformable image registration with a diffusion regularization on the spatial gradients of spatial transformation $\psi$.
- Diffeomorphic Voxelmorph (DifVM) [12] - a Voxelmorph-based network for probabilistic diffeomorphic registration, which guarantees diffeomorphic (topology-preserving) spatial transformations.
- Laplacian Pyramid Image Registration Network (LapIRN) [55] – a Laplacian pyramid network for deformable image registration, which was the top-performing method at the 2021 Learn2Reg challenge [56].

We inserted our AAN into the above DLRs, and then compared the AAN-enhanced DLRs (AAN-VM/DifVM/LapIRN) with their non-AAN counterparts (VM/DifVM/LapIRN) to evaluate the effect of AAN on registration performance.

We also compared the AAN-enhanced DLRs to three state-of-the-art traditional ORs:

- FNIRT [19] – a registration tool based on a weighted sum of scaled sum-of-squared differences and membrane energy.
- SyN [20] – a symmetric normalization method for maximizing the cross-correlation within the space of diffeomorphic maps, which is the top-performing method in the comparison study carried out by Klein et al [21].
- LCC-Demons [30] - A fast, robust registration method based on the log-Demons diffeomorphic registration algorithm. The transformation is parameterized by stationary velocity fields, and the similarity metric implements a symmetric local correlation coefficient.

### 4.3. Evaluation Metrics

We followed a commonly used evaluation scheme where a Dice metric is used to evaluate the registration performance [3][4][7][12]. Specifically, we warp the moving image and its segmentation mask, and then use the Dice metric to measure the overlap between the warped segmentation mask of the moving image and the segmentation mask of the fixed image. In general, a higher Dice score indicates a better registration result. For statistical analysis, we performed a paired Student's t-test, and a two-sided $P$ value < 0.05 is considered to indicate a statistically significant difference.

The smoothness and invertibility of the predicted spatial transformation $\psi$ were evaluated using the number of negative Jacobian determinants. Specifically, the Jacobian matrix $J_\psi(p)$ captures the local properties of $\psi$ around voxel $p$. The $\psi$ is invertible only at voxels where $|J_\psi(p)| > 0$ [47]. We counted the total number of voxels where $|J_\psi(p)| \leq 0$ for evaluation. Generally, a lower number of negative Jacobian determinants indicates a smoother and more invertible $\psi$.

### 4.4. Experimental Settings

We first experimented with subject-to-subject registration, in which we trained the network using random pairs of training images as input, and tested registration between pairs of unseen testing images. 50, 100, and 200 pairs of testing images were randomly chosen from IBSR18, Mindboggle101, and LPBA40 testing sets for evaluation.

Then, we performed atlas-based registration experiments, where we fine-tuned the trained network to register all images with an atlas image. Atlas-based registration is a common formulation in population analysis, where inter-subject registration is a core problem [7]. Normally, the atlas is an average image and is usually constructed by jointly and repeatedly aligning a dataset of brain MRI volumes and averaging them together [7]. We did not develop an atlas but used an existing atlas, the MNI-152 brain template [53]. We registered all testing images with the atlas and evaluated the performance using pairwise Dice among all testing subjects.

We also evaluated our AAN from the following aspects: (i) we measured the smoothness and invertibility of the predicted spatial transformations, (ii) we measured the additional runtime required for using AAN, (iii) we compared the training convergence of AAN-enhanced DLRs with their non-AAN counterparts, (iv) we analyzed the effect of training set size on registration performance, (v) we explored the effect of edge map quality on registration performance and, (vi) we performed a hyper-parameter analysis for the regularization parameter $\lambda$ of $\mathcal{L}_{structure}$.

In the analysis of training set size (iv), we also compared the AAN-VM with the VM enhanced by data argumentation. The data augmentation was performed on image appearance using the Appearance Transform Model (ATM) proposed by Zhao et al. [16]. The ATM can change image appearance and was used to augment the appearance variations for the training samples during training.

### 4.5. Implementation Details

We conducted standard brain MRI preprocessing steps for each scan, including intensity normalization, brain extraction, histogram matching, and affine transformation, by FreeSurfer [23] and FLIRT [24]. All images were affine-transformed and resampled to align with the MNI-152 brain template [53] with 1mm isotropic voxels, and then cropped to 144×192×160. For automatic edge extraction, we adopted 3D canny edge detection [36], in which a double threshold scheme is used to distinguish main edges from false edges caused by noise and intensity variations. We fixed the high threshold (HT) to be 0.2 and then varied the value of low threshold (LT) to control the edge map quality. The baseline ORs were implemented using publicly available software/packages with the hyper-parameters fine-tuned on the validation sets.

We implemented all models using Keras with a Tensorflow backend [26] on either a 12GB NVIDIA Titan X GPU or a 12 GB NVIDIA Titan V GPU, and we did not notice any appreciable differences between the different GPUs for training. To reduce memory usage, each training batch consists of one pair of fixed/moving volumes. We trained our networks with 100,000 iterations in a subject-to-subject registration fashion, where for each iteration two MRI volumes were randomly picked up from the training set to be the fixed and moving images. In the atlas-based registration experiment, we further fine-tuned the trained networks with 10,000 iterations using an atlas as the fixed image. It took approximately 18, 16, and 23 minutes to train VM, DifVM, and LapIRN for 1000 iterations on a NVIDIA Titan V GPU. AAN-enhanced DLRs, when compared to their non-AAN counterparts, required approximately 6 additional minutes for training. We used ADAM optimizer [27] with a learning rate of $10^{-4}$ for subject-to-subject training and $10^{-5}$ for atlas-based fine-tuning. All hyper-parameters are decided based on the performance on three validation sets. Our code is publicly available at *https://github.com/MungoMeng/Registration-AAN*.

## 5. Results

### 5.1. Subject-to-Subject Registration

The average Dice scores over all testing image pairs and all labeled anatomical structures are shown in Table I for subject-to-subject registration on three datasets. We used MSE (Eq. 7) or LCC (Eq. 9) as the $\mathcal{L}_{sim}$ and reported the results separately. The results show that AAN-enhanced DLRs achieved significantly higher Dice than their non-AAN counterparts. Compared to the baseline ORs, AAN-enhanced DLRs achieved higher or competitive results on all datasets. The LapIRN achieved the highest Dice among all baseline methods, and the AAN-LapIRN achieved the highest Dice among all registration methods.

Table I. Results (Dice) for subject-to-subject registration

| Method | IBSR18 | Mindboggle101 | LPBA40 |
|---|---|---|---|
| Before Registration | 0.600 | 0.355 | 0.624 |
| FNIRT | 0.697 | 0.464 | 0.645 |
| SyN | 0.716 | 0.535 | 0.703 |
| LCC-Demons | 0.704 | 0.514 | 0.686 |
| VM (MSE) | 0.694 | 0.516 | 0.684 |
| **AAN-VM (MSE)** | **0.718 (+0.024) ‡** | **0.545 (+0.029) ‡** | **0.702 (+0.018) ‡** |
| VM (LCC) | 0.711 | 0.531 | 0.689 |
| **AAN-VM (LCC)** | **0.727 (+0.016) ‡** | **0.551 (+0.020) ‡** | **0.703 (+0.014) ‡** |
| DifVM (MSE) | 0.698 | 0.478 | 0.688 |
| **AAN-DifVM (MSE)** | **0.716 (+0.018) ‡** | **0.512 (+0.034) ‡** | **0.701 (+0.013) ‡** |
| DifVM (LCC) | 0.712 | 0.486 | 0.694 |
| **AAN-DifVM (LCC)** | **0.724 (+0.012) ‡** | **0.515 (+0.029) ‡** | **0.704 (+0.010) ‡** |
| LapIRN (MSE) | 0.725 | 0.568 | 0.718 |
| **AAN-LapIRN (MSE)** | **0.746 (+0.021) ‡** | **0.592 (+0.024) ‡** | **0.729 (+0.11) ‡** |
| LapIRN (LCC) | 0.736 | 0.582 | 0.722 |
| **AAN-LapIRN (LCC)** | **0.748 (+0.012) ‡** | **0.595 (+0.013) ‡** | **0.730 (+0.08) ‡** |

**Bold**: The results of AAN-enhanced DLRs are in bold with the Dice improvements over their non-AAN counterparts in parentheses.
‡: $P < 0.05$, in comparison to its non-AAN counterpart.

### 5.2. Atlas-based Registration

The average Dice scores over all testing subjects and all labeled anatomical structures are shown in Table II for atlas-based registration on three datasets. The results indicate that AAN-enhanced DLRs also achieved significantly higher Dice than their non-AAN counterparts. Furthermore, compared to subject-to-subject registration, AAN-enhanced DLRs achieved higher results and surpassed the baseline ORs by a larger margin on the atlas-based registration. The LapIRN also achieved the highest Dice among all baseline methods, and the AAN-LapIRN achieved the highest Dice among all registration methods.

Table II. Results (Dice) for atlas-based registration

| Method | IBSR18 | Mindboggle101 | LPBA40 |
|---|---|---|---|
| Before Registration | 0.599 | 0.353 | 0.622 |
| FNIRT | 0.695 | 0.466 | 0.646 |
| SyN | 0.714 | 0.532 | 0.702 |
| LCC-Demons | 0.702 | 0.516 | 0.685 |
| VM (MSE) | 0.709 | 0.532 | 0.696 |
| **AAN-VM (MSE)** | **0.734 (+0.025) ‡** | **0.558 (+0.026) ‡** | **0.708 (+0.012) ‡** |
| VM (LCC) | 0.720 | 0.545 | 0.698 |
| **AAN-VM (LCC)** | **0.736 (+0.016) ‡** | **0.565 (+0.020) ‡** | **0.707 (+0.009) ‡** |
| DifVM (MSE) | 0.705 | 0.501 | 0.698 |
| **AAN-DifVM (MSE)** | **0.725 (+0.020) ‡** | **0.533 (+0.032) ‡** | **0.708 (+0.010) ‡** |
| DifVM (LCC) | 0.721 | 0.504 | 0.700 |
| **AAN-DifVM (LCC)** | **0.733 (+0.012) ‡** | **0.532(+0.028) ‡** | **0.708 (+0.008) ‡** |
| LapIRN (MSE) | 0.732 | 0.575 | 0.723 |
| **AAN-LapIRN (MSE)** | **0.751 (+0.019) ‡** | **0.596 (+0.021) ‡** | **0.733 (+0.10) ‡** |
| LapIRN (LCC) | 0.741 | 0.588 | 0.726 |
| **AAN-LapIRN (LCC)** | **0.752 (+0.011) ‡** | **0.599 (+0.011) ‡** | **0.733 (+0.07) ‡** |

**Bold**: The results of AAN-enhanced DLRs are in bold with the Dice improvements over their non-AAN counterparts in parentheses.
‡: $P < 0.05$, in comparison to its non-AAN counterpart.

### 5.3. Transformation Smoothness and Invertibility

The average number of voxels with negative Jacobian determinants over all testing subjects is shown in Table III. Table I and Table II shows that our AAN consistently improved DLRs on the subject-to-subject and atlas-based registration, and therefore we only report the results of subject-to-subject registration in Table III and in the following experiments. The results show that the spatial transformations derived from AAN-enhanced DLRs exhibit relatively consistent smoothness and invertibility with the ones derived from their non-AAN counterparts. Examples of Jacobian determinant maps are shown in Fig. 4, which also shows that the AAN does not impact on the smoothness and invertibility of spatial transformations.

Table III. The average number of negative Jacobian determinants

| Method | IBSR18 | Mindboggle101 | LPBA40 |
|---|---|---|---|
| VM (MSE) | 32257 | 94568 | 13425 |
| **AAN-VM (MSE)** | **28875 (-3382)** | **88369 (-6199)** | **10248 (-3177)** |
| VM (LCC) | 46756 | 105263 | 35779 |
| **AAN-VM (LCC)** | **44836 (-1920)** | **99864 (-5399)** | **32649 (-3130)** |
| DifVM (MSE) | 215 | 268 | 128 |
| **AAN-DifVM (MSE)** | **167 (-48)** | **197 (-71)** | **86.8 (-41.2)** |
| DifVM (LCC) | 87.6 | 112 | 29.6 |
| **AAN-DifVM (LCC)** | **128 (+40.4)** | **249 (+137)** | **67. 4 (+37.8)** |
| LapIRN (MSE) | 22863 | 105123 | 11004 |
| **AAN-LapIRN (MSE)** | **20695 (-2168)** | **99005 (-6118)** | **8397 (-2607)** |
| LapIRN (LCC) | 41656 | 93929 | 33174 |
| **AAN-LapIRN (LCC)** | **39664 (-1992)** | **88592 (-5337)** | **29618 (-3556)** |

**Bold**: The results of AAN-enhanced DLRs are in bold with the differences caused by AAN in parentheses.

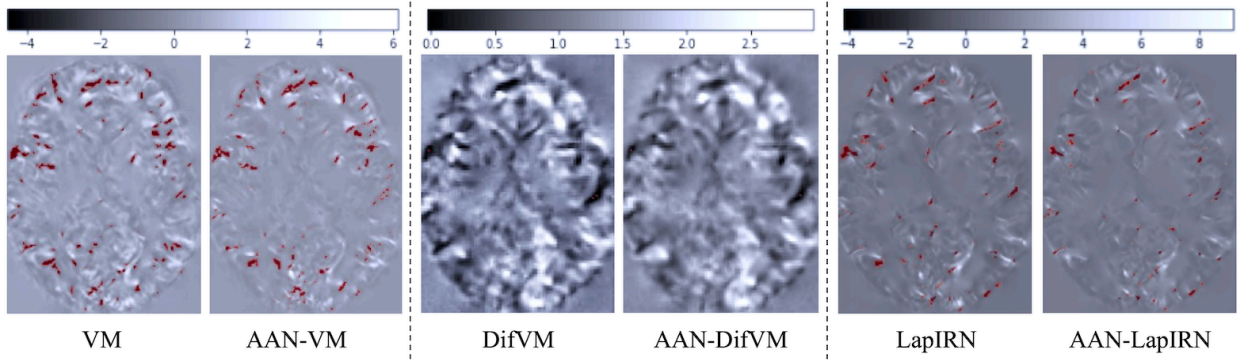

Fig. 4. A visual comparison of the Jacobian determinant maps produced by AAN-enhanced DLRs (AAN-VM/DifVM/LapIRN) and their non-AAN counterparts (VM/DifVM/LapIRN). The locations with negative Jocobian determinants are highlighted in red.

### 5.4. Runtime

The average runtime for registering a pair of images is listed in Table IV, which was measured using an Intel Core i5-9400 CPU or a NVIDIA Titan V GPU. The runtime results of three baseline ORs are also listed for comparison. Note that since the GPU implementations of the baseline ORs are not currently available, we only ran them on the CPU. The results show that the runtime of AAN-enhanced DLRs, compared to their non-AAN counterparts, increased by nearly 1.2/0.1 second on the CPU/GPU. Nevertheless, the runtime of AAN-enhanced DLRs is still much shorter than that of the baseline ORs. FNIRT/LCC-Demons required approximately 13/10 minutes and SyN required more than 60 minutes on the CPU, while AAN-enhanced DLRs can register a pair of images within 10 seconds on the CPU and within a second on the GPU.

Table IV. The average runtime for registration after the preprocessing steps

| Method | CPU (second) | GPU (second) |
|---|---|---|
| FNIRT | 786 | \ |
| SyN | 3816 | \ |
| LCC-Demons | 634 | \ |
| VM | 2.25 | 0.28 |
| **AAN-VM** | **3.45 (+1.20)** | **0.36 (+0.08)** |
| DifVM | 1.92 | 0.27 |
| **AAN-DifVM** | **3.18 (+1.26)** | **0.35 (+0.08)** |
| LapIRN | 5.02 | 0.48 |
| **AAN-LapIRN** | **6.31 (+1.29)** | **0.57 (+0.09)** |

**Bold**: The results of AAN-enhanced DLRs are in bold with the additional runtime required for AAN in parentheses.

### 5.5. Training Convergence

We compared the training convergence of AAN-enhanced DLRs to their non-AAN counterparts, and the results are shown in Fig. 5. Since both the MSE and LCC losses showed similar behavior, we reported MSE here and in the experiments following. The MSE training loss, with varying number of training iterations, is plotted in Fig. 5. The training curves show that the AAN-enhanced DLRs have faster training convergence and achieve lower training loss than their non-AAN counterparts. It should be noted that, the LapIRN and the AAN-LapIRN were trained using a multi-stage training strategy [55], and we only show the training curve at the first stage in Fig. 5c for illustration.

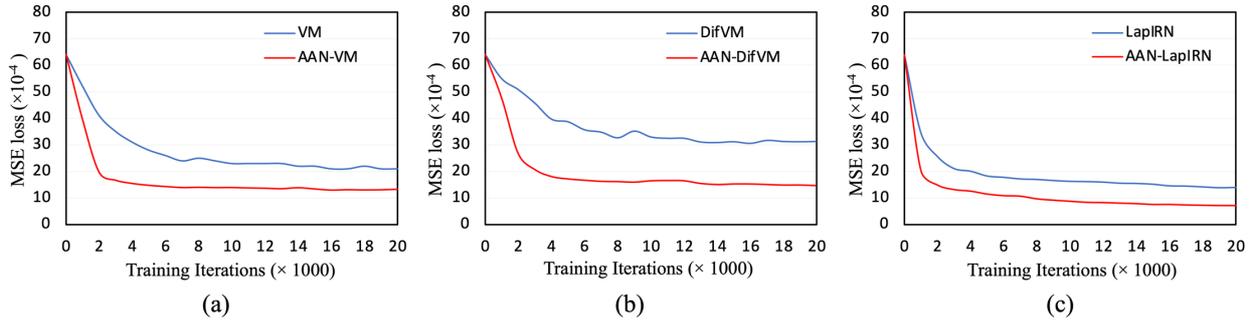

Fig. 5. Comparison of training convergence between (a) VM and AAN-VM, (b) DifVM and AAN-DifVM, and (c) LapIRN and AAN-LapIRN.

### 5.6. Training Set Size Analysis

We trained AAN-VM and VM using varying numbers of training samples in the training set, and the Dice scores on the IBSR18 validation set are shown in Fig. 6. The results show that our AAN consistently improved the VM and showed a greater improvement when a smaller number of training samples are available. The data augmentation can also improve the VM, but the improvements become marginal when the training samples are sufficient.

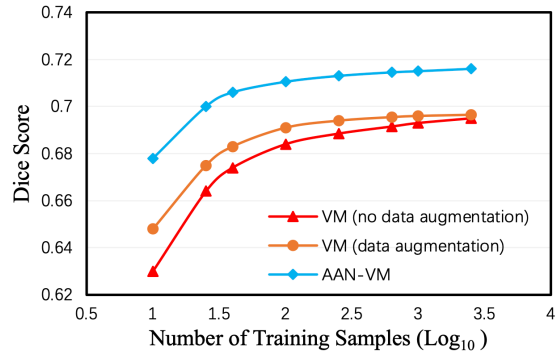

Fig. 6. Registration results (Dice) produced with varying numbers of training samples.

### 5.7. Edge Extraction Analysis

We analyzed the effect of edge map quality on the registration performance. A visualization of the edge maps produced by Canny edge detection with different threshold settings is shown in Fig. 7. Fig. 8 shows the average Dice scores on the IBSR18 validation set with varying low threshold (LT), and the high threshold (HT) is fixed as 0.2. As is shown in Fig.8, the results of AAN-VM consistently outperformed the VM.

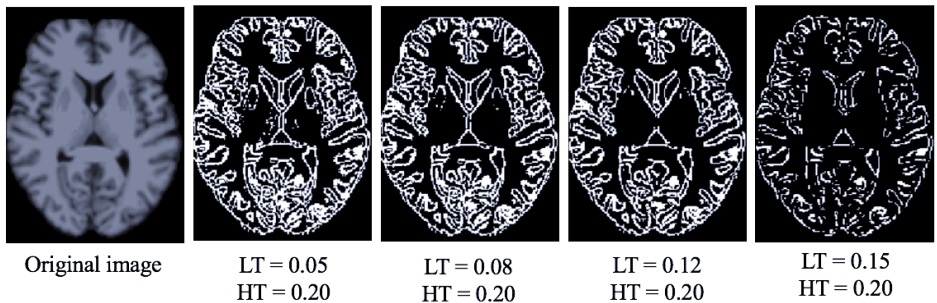

Fig. 7. A visualization of the edge maps produced by Canny edge detection with varying double threshold settings.

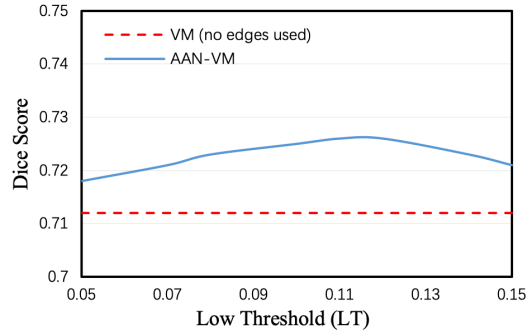

Fig. 8. Effect of varying edge detection parameter LT on Dice Score.

## 5.8. Regularization Analysis

We performed hyper-parameter analysis for the regularization parameter $\lambda$ of $\mathcal{L}_{structure}$. Fig. 9 shows the MSE training loss and the average Dice scores on the IBSR18 validation set with varying values of parameter $\lambda$. When the $\lambda$ becomes too large, the Dice score of AAN-VM degrades and approaches the result of VM. When the $\lambda$ becomes too small, the $\mathcal{L}_{sim}$ of AAN-VM tends to decrease but its Dice score also drops below VM. We performed the same analyses on the LPBA40 and Mindboggle101 validation sets and observed similar behavior as shown in Fig. 9. A visualization of the appearance transformations produced with varying $\lambda$ is plotted in Fig. 10. As shown in Fig. 10, too large $\lambda$ values make appearance transformations lack local variations, while too small $\lambda$ values result in the appearance transformations that potentially alter the original anatomical structures.

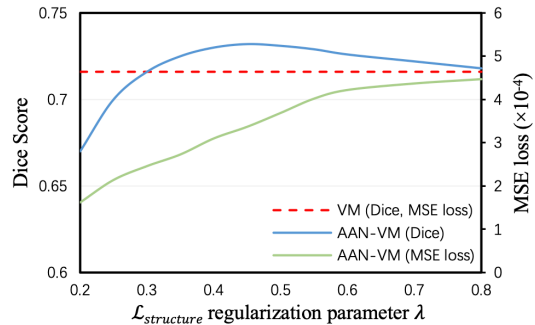

Fig. 9. Effect of varying $\mathcal{L}_{structure}$ regularization parameter $\lambda$ on Dice Score and MSE $\mathcal{L}_{sim}$ value. When $\lambda = 0.45$, the AAN-VM achieves the highest Dice Score.

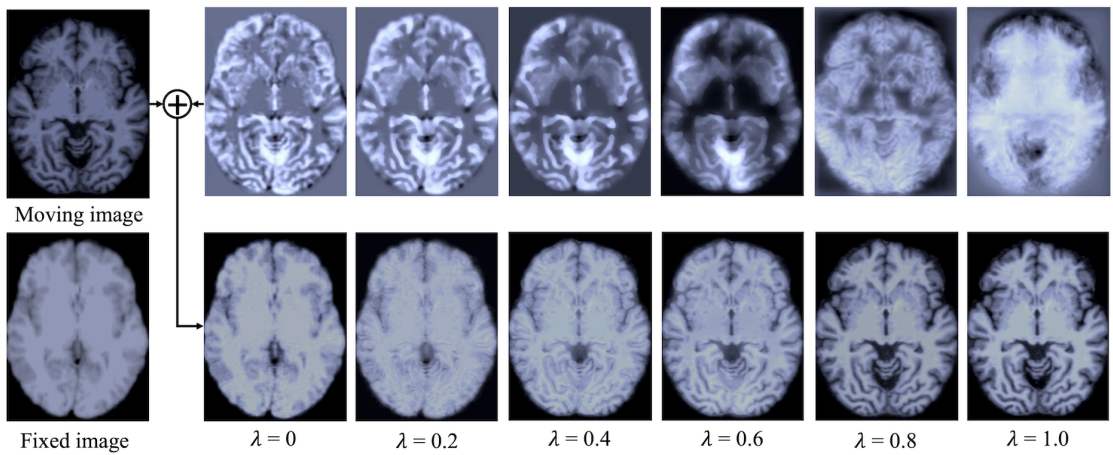

Fig. 10. A visualization of the appearance transformations produced with varying regularization parameter $\lambda$.

# 6. Discussion

Our main findings are: (i) our AAN improved the registration accuracy of DLRs while not degrading the transformation smoothness and only adding a fractional computational load , (ii) our AAN facilitated the training convergence of DLRs and reduced the requirement for the number of training samples, (iii) our AAN had stable performance regardless of the quality of automatically-extracted edge maps, and (iv) the regularization parameter $\lambda$ can balance adjusting appearance and preserving anatomy, which also aids in explaining the effects of our AAN on DLRs.

We found that AAN-enhanced DLRs significantly improved Dice scores over non-AAN counterparts on the subject-to-subject registration (Table I) and atlas-based registration (Table II). The LapIRN achieved the highest Dice among all baseline methods (FNIRT, SyN, LCC-Demons, VM, and DifVM), which is consistent with the results reported by Mok et al. [55]. This can be attributed to the fact that the LapIRN decoupled the registration process with 3-level pyramid networks and optimized the registration in a coarse-to-fine manner. Nevertheless, the LapIRN did not take into account appearance variations and can be further improved by our AAN. Consequently, the AAN-LapIRN achieved the highest Dice among all methods. It is important to note that with Dice scores, better registration accuracy potentially comes at the expense of smooth and invertible transformations [3][46]. Nevertheless, we identified that our AAN does not degrade the smoothness and invertibility of spatial transformations (Table III and Fig. 4) because the number of negative Jacobian determinants is still determined by the baseline DLRs (VM, DifVM, or LapIRN) and the loss functions (MSE or LCC). Furthermore, the runtime results (Table IV) show that our AAN only adds a fractional computational load to the existing DLRs, and thus still retains an advantage over ORs.

The Dice results (Table I and Table II) show that the DLRs can achieve better performance on the atlas-based registration than subject-to-subject registration. We believe that this can be explained by less variability on the atlas-based registration, which is consistent with the results from Balakrishnan et al. [3]. In contrast to DLRs, all the baseline ORs exhibited similar results on the subject-to-subject and atlas-based registration. We attribute this to the ORs' target-pair-specific optimization, which is optimized for individual pairs, regardless of subject-to-subject registration or atlas-based registration. Consequently, compared to subject-to-subject registration, AAN-enhanced DLRs outperformed the baseline ORs by a larger margin on the atlas-based registration. Moreover, compared to IBSR18 and LPBA40, there are larger appearance variations within Mindboggle101, as the Mindboggle101 dataset consists of MRI scans acquired from different source datasets (e.g., OASIS, MMRR, NKI, etc.). We suggest that this explains the greater improvement of AAN-enhanced DLRs over non-AAN counterparts on this dataset.

We also experimented with two widely used loss functions, MSE (Eq. 7) and LCC (Eq. 9), for $\mathcal{L}_{sim}$. The LCC loss is considered more robust to appearance variations than the MSE loss [20], which is consistent with our results that the LCC loss contributed to better performance than the MSE loss in the baseline DLRs (Table I and Table II). Our results also showed that the LCC loss cannot fully address appearance variations, and our AAN can still improve DLRs even when the LCC loss has already been used (Table I and Table II). The LCC loss and our AAN address appearance variations differently: the LCC loss improves the robustness to appearance variations, while our AAN reduces the appearance variations in both training and testing samples. Therefore, the AAN-enhanced DLRs can be further improved by the LCC loss. Nevertheless, it should be noted that, for AAN-enhanced DLRs, the improvements caused by the LCC loss is marginal as our AAN has removed the most appearance variations.

In the comparison of training convergence, we found that AAN-enhanced DLRs exhibited faster training convergence than their non-AAN counterparts (Fig. 5). We suggest that, with the assistance of our AAN, DLRs can focus on learning spatial transformation rather than being distracted by appearance variations, which results in a faster decline in the training loss. We also found that AAN-enhanced DLRs had a lower training loss when fully trained (Fig. 5). A lower training loss, however, does not always indicate better performance and this can be attributed to the appearance adjusted moving image having higher intensity similarity with the fixed image, which naturally results in lower training loss.

In the analysis of training set size, we found that the data augmentation on image appearance can improve DLRs, but the improvements are marginal when the training samples were sufficient. This is likely because, when the training samples are sufficient, the appearance variations in training samples can well resemble the variations in testing samples. In this case, the registration performance is bounded by the network capability and the data augmentation has a limited effect on improving the performance. In contrast, our AAN reduces the appearance variations in both training and testing samples, which enables DLRs to allocate more network capability to learn spatial variations. Therefore, the AAN-VM can consistently outperform the VM, regardless of whether training samples are sufficient or insufficient. It should be noted that data augmentation on image appearance cannot be used together with the AAN as they perform opposite operations on appearance variations. We also found that the AAN-VM exhibited greater Dice improvements over the VM when limited training data were available (Fig. 6), which suggests that the AAN-VM could be trained with less training data than the VM. We attribute this to the AAN-enhanced DLRs needing to learn less appearance variations and thus being able to focus on learning spatial transformations.

In the analysis of edge extraction, the results of AAN-VM consistently exceeded VM and varied consistently over a large range of LT values (Fig. 8), illustrating that the AAN performs consistently regardless of the quality of edge maps. A clean edge map containing only main edges is desirable on edge detection tasks, but it is difficult to obtain an ideally clean edge map through automatic algorithms. Nonetheless, with coarse edge maps (such as in Fig. 7) with false positive edges or where some main edges are lost, our AAN still contributed to better registration results. The AAN-VM achieved the highest result when LT≈0.12, but the edge map with LT=0.12 (in Fig. 7) is still far from being ideally clean. We anticipate that a more accurate edge map will result in a more accurate registration. We will explore this issue using more accurate edge maps in our future study.

In regard to the analysis of the regularization parameter $\lambda$, when the $\lambda$ reached high values, the Dice score of AAN-VM degraded and approached that of VM (Fig.9). This is expected as the appearance adjustment is being restricted excessively, and thus lacks local variations and tends to act as a global intensity adjustment (Fig.10). In contrast, when the $\lambda$ was set to too small values, the $\mathcal{L}_{sim}$ tended to decrease but the Dice score also dropped (Fig.9). In this context, the appearance adjustment does not derive sufficient restriction, which results in the changes to the original anatomical structures of the moving image (Fig.10). This experiment implicitly shows the effects of our AAN on baseline DLRs, where the AAN with an appropriate $\lambda$ can adjust the image's appearance without changing its original anatomical structures and this helps to reduce appearance variations and enhances registration accuracy. Further, we observed that the choice of $\lambda$ is robust to different datasets. Therefore, we did not optimize different $\lambda$ values for different datasets, instead we used the same $\lambda$ value to train models and then tested them on three testing sets.

In this study, we used brain MRI images but our AAN is not restricted to these images and can be applied to other body organs, such as lungs for motion corrections, or other imaging modalities, such as Positron Emission Tomography – Computed Tomography (PET-CT). In addition, we focused on inserting our AAN into unsupervised DLRs, but we expect that supervised or weakly-supervised DLRs also can benefit from our AAN because appearance variations are also undesirable in these settings. We plan to investigate the integration of AAN into supervised or weakly-supervised DLRs in our future study.

## 7. Conclusion

We propose an appearance adjustment network to enhance deep learning-based registration methods through reducing appearance variations. Our appearance adjustment network enhances the adaptability of deep learning-based registration methods to appearance variations and improves their learning efficiency. Further, our anatomy-constrained loss function based on anatomy edge maps restrains anatomical deformations that can occur during appearance transformations. Our results with three state-of-the-art unsupervised deep learning-based registration methods and three public brain MRI datasets demonstrate that our appearance adjustment network improves image registration with a marginal additional computational load.